\pdfoutput=1

\documentclass{article}

\usepackage{microtype}
\usepackage{graphicx}

\usepackage{acronym}
\usepackage{xstring}
\usepackage[hyphens]{url}
\usepackage{xcolor}
\usepackage{graphicx}
\usepackage{tikz}

\usepackage{amsmath}
\usepackage{amssymb}

\usepackage{placeins}

\usepackage{booktabs} 
\usepackage{array}
\usepackage{subcaption}
\usepackage{multirow}
\usepackage{hhline}
\usepackage{colortbl}
\usepackage{collcell}
\usepackage{stringstrings}

\usepackage{hyperref}
\hypersetup{colorlinks=true,breaklinks=true}

\newcommand\rurl[1]{%
  \href{http://#1}{\nolinkurl{#1}}%
}

\newcommand*\rot{\rotatebox{90}}%
\newcommand*{\MinNumber}{-1}%
\newcommand*{\MidNumber}{0}%
\newcommand*{\MaxNumber}{1}%

\newcommand{\ApplyGradient}[1]{%
    \getargs[q]{#1}
    \ifdim \narg pt > 0 pt
        \edef\argnumber{\argi}
        \IfDecimal{\argnumber}{
            \ifdim \argnumber pt > \MidNumber pt
                \pgfmathsetmacro{\PercentColor}{max(min(100.0*(\argnumber - \MidNumber)/(\MaxNumber-\MidNumber),100.0),0.00)}
    
                \ifdim \narg pt < 2 pt
                    \edef\x{ \noexpand\cellcolor{green!\PercentColor!yellow} }
                \else
                    \edef\x{ \noexpand\cellcolor{\argii} }
                \fi
                \x \argnumber
            \else
                \pgfmathsetmacro{\PercentColor}{max(min(100.0*(\MidNumber - \argnumber)/(\MidNumber-\MinNumber),100.0),0.00)}
                
                \ifdim \narg pt < 2 pt
                    \edef\x{ \noexpand\cellcolor{green!\PercentColor!yellow} }
                \else
                    \edef\x{ \noexpand\cellcolor{\argii} }
                \fi
                \x \argnumber
            \fi
        }{\argnumber}
    \else
    \fi
}
\newcolumntype{R}{>{\collectcell\ApplyGradient}r<{\endcollectcell}}

\usepackage[accepted]{icml2021}

\acrodef{XAI}{Additive Explainable AI}
\acrodef{NLP}{Natural Language Processing}

\begin{document}


\twocolumn[
\icmltitle{Order in the Court: Explainable AI Methods Prone to Disagreement}



\icmlsetsymbol{equal}{*}

\begin{icmlauthorlist}
\icmlauthor{Michael Neely}{equal,uva}
\icmlauthor{Stefan F. Schouten}{equal,uva}
\icmlauthor{Maurits J. R. Bleeker}{uva}
\icmlauthor{Ana Lucic}{uva}
\end{icmlauthorlist}

\icmlaffiliation{uva}{University of Amsterdam}

\icmlcorrespondingauthor{Michael Neely}{michael@mneely.tech}
\icmlcorrespondingauthor{Stefan F. Schouten}{sfschouten@gmail.com}
\icmlcorrespondingauthor{Maurits J. R. Bleeker}{m.j.r.bleeker@uva.nl}
\icmlcorrespondingauthor{Ana Lucic}{a.lucic@uva.nl}

\icmlkeywords{Machine Learning, ICML}

\vskip 0.3in
]



\printAffiliationsAndNotice{\icmlEqualContribution} 
\setcounter{footnote}{1}

\begin{abstract}
By computing the rank correlation between attention weights and feature-additive explanation methods, previous analyses either invalidate or support the role of attention-based explanations as a faithful and plausible measure of salience. To investigate whether this approach is appropriate, we compare LIME, Integrated Gradients, DeepLIFT, Grad-SHAP, Deep-SHAP, and attention-based explanations, applied to two neural architectures trained on single- and pair-sequence language tasks. In most cases, we find that none of our chosen methods agree. Based on our empirical observations and theoretical objections, we conclude that rank correlation does not measure the quality of feature-additive methods. Practitioners should instead use the numerous and rigorous diagnostic methods proposed by the community. 
\end{abstract}

\section{Introduction} \label{sec:introduction}

Of the many possible explanations for a model's decision, only those simultaneously \emph{plausible} to human stakeholders and \emph{faithful} to the model's reasoning process are desirable \cite{jacovi-goldberg-2020-towards}. The rest are irrelevant in the best case and harmful in the worst, particularly in critical domains such as law \cite{Kehl2017AlgorithmsIT}, finance \cite{grath2018interpretable}, and medicine \cite{caruana2015intelligble}. It would therefore be prudent to discourage algorithms that generate misleading explanations. However, it is challenging to identify when \acf{XAI} methods fail without first decomposing the abstract concepts of plausibility and faithfulness into measurable diagnostic properties.

In their critique of attention-based explanations, \citet{jain-wallace-2019-attention} argue that faithful \ac{XAI} methods\footnote{For the sake of brevity, we refer to all feature-additive algorithms (e.g. \citeauthor{lime}, \citeyear{lime}) simply as `\ac{XAI} methods'.} must be highly \emph{agreeable}\footnote{\citet{ethayarajh2021attention} use the term \emph{consistent}.}. That is, their generated rankings of input importance must correlate with other \ac{XAI} methods. Following \citet{jain-wallace-2019-attention}'s claim that `attention is not explanation', several recent papers have presented an increased agreement with a small set of \ac{XAI} methods as evidence for their proposed method's ability to improve the faithfulness of the attention mechanism. For example, \citet{mohankumar-etal-2020-towards} show that minimizing hidden state conicity in a BiLSTM improves the Pearson correlation of attention weights with Integrated Gradients \cite{intgrad} attributions. As the popularity of \emph{agreement as evaluation} grows \citep[][\textit{inter alia}]{meister2021sparse,abnar-zuidema-2020-quantifying}, we believe it is worth investigating the diagnostic capacity of agreement as a metric.

Under the paradigm of \emph{agreement as evaluation}, proposed \ac{XAI} methods (e.g., attention-based) are compared to one or more established \ac{XAI} method(s) (e.g., gradient-based). 
However, can any \ac{XAI} method act as the standard against which other \ac{XAI} methods may be graded?
Explanations are task-, model-, and context-specific \cite{doshivelez2017rigorous}, and the performance of \ac{XAI} methods depends on the particular diagnostic tests considered \citep[][\textit{inter alia}]{deyoung-etal-2020-eraser,Robnik-sikonja2018}. 
In this work, we examine the agreement of contemporary \ac{XAI} methods in a more expansive study to investigate what \textit{agreement as evaluation} can lead us to conclude. We ask:

\noindent\textit{\textbf{RQ:} How well do the \ac{XAI} methods LIME, Integrated Gradients, DeepLIFT, Grad-SHAP, and Deep-SHAP correlate \textbf{(i)} with one other and \textbf{(ii)} with attention-based explanations? Does the correlation depend on \textbf{(a)} the model architecture (LSTM- and Transformer-based), or \textbf{(b)} the nature of the classification task (single- and pair-sequence)? }

We observe low overall agreement between XAI methods, particularly for the more complex Transformer-based model, and pair-sequence tasks. We use this empirical evidence, along with our theoretical objections, to argue that practitioners should refrain from grading \ac{XAI} methods based on agreement. Rank correlation is not a method of objective evaluation unless ground-truth 
rankings are available \citep[e.g., in ][]{yalcin2021evaluating}. In all other situations, rigorous diagnostic measures — such as those proposed by \citet{atanasova-etal-2020-diagnostic} — are better suited for this role.

\section{Related Work}

\subsection{Agreement as Evaluation}
\citet{jain-wallace-2019-attention} introduced the \emph{agreement as evaluation} paradigm by comparing attention-based explanations with simple \ac{XAI} methods. Specifically, they report a weak Kendall-$\tau$ correlation between the rankings of input token importance obtained from attention weights and those from the input $\times$ gradient \citep{inputxgradients1, inputxgradients2} and leave-one-out \cite{li2017understanding} \ac{XAI} methods. Their work inspired others to measure agreement, including \citet{abnar-zuidema-2020-quantifying}, who demonstrate their \emph{attention-flow} algorithm improves the SpearmanR correlation with the feature-ablation (blank-out) XAI method, and \citet{meister2021sparse}, who show that — under the same experimental setup as \citet{jain-wallace-2019-attention} — inducing sparsity in the attention distribution decreases agreement with \ac{XAI} methods. We test the generalizability of agreement as a metric by including a more complex Transformer-based model and by comparing more recent \ac{XAI} methods.

\subsection{Attention as Explanation}
Despite concerns with \citet{jain-wallace-2019-attention}'s approach \citep[][\textit{inter alia}]{wiegreffe-pinter-2019-attention, grimsley-etal-2020-attention}, their influential critique has inspired enhancements of the faithfulness and plausibility of attention-based explanations. Proposed modifications of the attention mechanism include: guided training \cite{zhong2019finegrained}, sparsity \cite{correia-etal-2019-adaptively}, and word-level objectives \cite{tutek-snajder-2020-staying}. Additionally, techniques such as projecting from the null space of multi-head self-attention \cite{brunner2019identifiability}, or accounting for the transformed vectors' magnitude \cite{kobayashi-etal-2020-attention}, address problems with analyzing attention weights in their raw form. \citet{bastings-filippova-2020-elephant} question why the community is concerned with the faithfulness of attention when salience measures already exist. 
We contribute to the `attention/explanation' argument by including attention-based explanations in our survey, but do not seek to justify their use.

\subsection{Limitations of \ac{XAI} methods}
\ac{XAI} methods are known to suffer from limitations. For example, \citet{camburu*2020can} show that LIME \cite{lime} and SHAP \cite{shap} tend to select a token with zero contribution as the most relevant feature, \citet{kindermans2019unreliability} show that saliency methods are not invariant to consistent transformations of model inputs, and \citet{NEURIPS2019_fe4b8556} demonstrate that gradient-based \ac{XAI} methods are no better than random rankings of importance under their remove-and-retrain approach. Finally, \citet{yalcin2021evaluating} prove the performance of TreeSHAP is inversely correlated with dataset complexity when ground-truth rankings of feature importance are known. \citet{atanasova-etal-2020-diagnostic} unify various evaluation paradigms with a series of diagnostic tests to evaluate \ac{XAI} methods for text classification. 
We also compare \ac{XAI} methods, but only to investigate the suitability of \textit{agreement as evaluation}.
\section{Method} \label{sec:method}

We define an \emph{explanation} of an input sequence of tokens as a vector of corresponding importance scores. We investigate two types of explanations: (i) those from recent \ac{XAI} methods and (ii) those based on attention scores. We measure \emph{agreement} between these explanation methods as the Kendall-$\tau$ correlation between the ranked importance scores of all input tokens.

\subsection{Recent XAI methods} \label{ssec:recent_methods}

We select a number of recent XAI methods, namely: {LIME}; {Integrated Gradients}; {DeepLIFT} \cite{deeplift}; and two methods from the {SHAP} family: {Grad-SHAP}, which is based on Integrated Gradients; and {Deep-SHAP}, which is based on DeepLIFT. 

\subsection{Attention-based explanations} \label{ssec:attn_methods}

Given an input sequence of tokens $S = t_{1}, ..., t_{n}$, we define an \emph{attention-based explanation} as an assignment of attention weights $\boldsymbol{\alpha} \in \mathbb{R}^{n}$ over the tokens in $S$. Since the dimensionality of $\boldsymbol{\alpha}$ is architecture-dependent, it may be necessary to filter or aggregate the weights. In our experiments, this is only relevant for our Transformer-based model's self-attention mechanism \cite{vaswani2017attention}. 
Previous analyses at the attention head level \citep[e.g.,][]{baan2019understanding, clark-etal-2019-bert} implicitly assume that contextual word embeddings remain tied to their corresponding tokens across self-attention layers. This assumption may not hold in Transformers, since information mixes across layers \cite{brunner2019identifiability}. Therefore, we use the \emph{attention rollout} \cite{abnar-zuidema-2020-quantifying} method --- which assumes the identities of tokens are linearly combined through the self-attention layers based exclusively on attention weights --- to calculate a post-hoc, faithful, token-level attribution. Like \citet{abnar-zuidema-2020-quantifying}, we use the attribution calculated for the last layer's [CLS] token, resulting in a final vector $\boldsymbol{\alpha} \in \mathbb{R}^{n}$ at the time of evaluation.

Recurrent models similarly suffer from issues of identifiability. In LSTM-based models, attention is computed over hidden representations across timesteps, which does not provide faithful token-level attribution. Approaches that trace explanations back to individual timesteps \cite{Bento2020TimeSHAPER} or input tokens \cite{tutek-snajder-2020-staying} are only just emerging. Therefore, we limit ourselves to an analysis of the raw attention weights for our LSTM-based model.

\section{Experiments} \label{sec:experiments}

\subsection{Datasets} \label{sec:datasets}
We evaluate two types of classification tasks: (i) single-sequence, and (ii) pair-sequence. For single-sequence, we perform binary sentiment classification on the Stanford Sentiment Treebank (\textbf{SST-2}) \cite{socher-etal-2013-parsing} and the \textbf{IMDb} Large Movie Reviews Corpus \cite{maas-etal-2011-learning}. We use identical splits and pre-processing as \citet{jain-wallace-2019-attention}, but also remove sequences longer than 240 tokens for faster attribution calculation. For pair-sequence, we examine natural language inference and understanding with the \textbf{SNLI} \cite{bowman-etal-2015-large}, \textbf{MultiNLI} \cite{N18-1101}, and \textbf{Quora} Question Pairs datasets. Since MultiNLI has no publicly available test set, we use the English subset of the XNLI \cite{conneau2018xnli} test set. We use a custom split  (80/10/10) for the Quora dataset, removing pairs with a combined count of 200 or more tokens. Most importantly, we include a uniform activation baseline to contextualize the attention mechanism's utility \cite{wiegreffe-pinter-2019-attention}.

\subsection{LSTM-based Model}
We use the same single-layered bidirectional encoder with additive $(tanh)$ attention and linear feedforward decoder as \citet{jain-wallace-2019-attention}. In pair-sequence tasks, we embed, encode, and induce attention over each sequence separately. The decoder predicts the label from the concatenation of: both context vectors $c_{1}$ and $c_{2}$; their absolute difference $|c_{1} - c_{2}|$; and their element-wise product $c_{1} \cdot c_{2}$.

\subsection{Transformer-based Model}
To reduce the computational overhead, we fine-tune the lighter, pre-trained DistilBERT variant \cite{sanh2019distilbert} instead of the full BERT model \cite{devlin-etal-2019-bert}. For classification, we add a linear layer on top of the pooled output. We concatenate pair-sequences with a [SEP] token.

\subsection{Training the models}
We train three independently-seeded instances of both models using the AllenNLP framework \cite{Gardner2017AllenNLP}, each for a maximum of 40 epochs. We use a patience value of 5 epochs for early stopping. For the BiLSTM, we follow \citet{jain-wallace-2019-attention} and select a 128-dimensional encoder hidden state with a 300-dimensional embedding layer. We tune pre-trained FastText embeddings \cite{bojanowski2016enriching} and optimize with the AMSGrad variant \cite{Tran_2019} of Adam \cite{kingma2017adam}. For DistilBERT, we fine-tune the standard `base-uncased' weights available in the HuggingFace library \cite{Wolf2019HuggingFacesTS} with the AdamW \cite{loshchilov2019decoupled} optimizer. 
Table \ref{tab:accuracy} confirms our models are sufficiently accurate for our analysis. 
Our extendable Python package for evaluating agreement between XAI methods and attention-based explanations, \texttt{court-of-xai}, is publicly available%
\footnote{\rurl{github.com/sfschouten/court-of-xai}}.

\subsection{Explaining the models}
We leverage existing implementations of LIME, Integrated Gradients, DeepLIFT, Grad-SHAP, and Deep-SHAP\footnote{\rurl{github.com/pytorch/captum}}, and use the padding token as a baseline where applicable. For LIME, we mask tokens as features and use 1000 samples to train the interpretable models. We apply our XAI methods to 500 random instances taken from each test set.

\begin{table}
    \centering
    \caption{Test set accuracy using softmax or uniform activations in the attention mechanisms. A uniform activation renders the mechanism defunct and contextualizes its utility for each task.}\label{tab:accuracy}
    \resizebox{\columnwidth}{!}{%
    \begin{tabular}{crrrr}
        \toprule
                & \multicolumn{2}{c}{BiLSTM} & \multicolumn{2}{c}{DistilBERT} \\ 
          \cmidrule(r){2-3}\cmidrule(r){4-5}
         & Uniform & Softmax & Uniform & Softmax\\
        \midrule
        MNLI    & $.659 \pm .001$   & $.667 \pm .004$   & $.599 \pm .002$   & $.779 \pm .002$ \\
        Quora   & $.829 \pm .001$   & $.830 \pm .001$   & $.832 \pm .001$   & $.888 \pm .001$ \\
        SNLI    & $.804 \pm .004$   & $.807 \pm .002$   & $.770 \pm .005$   & $.871 \pm .001$ \\
        IMDb    & $.874 \pm .011$   & $.872 \pm .014$   & $.879 \pm .003$   & $.890 \pm .005$ \\
        SST-2   & $.823 \pm .008$   & $.826 \pm .011$   & $.823 \pm .004$   & $.842 \pm .003$ \\
    \bottomrule
    \end{tabular}}
\end{table}
\section{Results}\label{sec:results}

\begin{table}
    \centering
    \caption{Mean Kendall-$\tau$ between the explanations given by our XAI methods for each model when applied to 500 instances of the test portion of each dataset. Comparisons between methods and their SHAP variants are not representative and thus colored gray.}\label{tab:results}
    \setlength\tabcolsep{2pt}
    \begin{subtable}[b]{0.97\columnwidth}%
        \renewcommand{\arraystretch}{1.1}%
        \resizebox{\columnwidth}{!}{%
        \begin{tabular}{clRRRRR}
            \toprule
            &  & \multicolumn{1}{c}{LIME} & \multicolumn{1}{c}{Int-Grad} & \multicolumn{1}{c}{DeepLIFT} & \multicolumn{1}{c}{Grad-SHAP}  & \multicolumn{1}{c}{Deep-SHAP} \\
            \hhline{-------}\addlinespace[0.1mm]
            \multirow{5}{*}{\rot{Attn}}
                & MNLI    & .1958  & .2523  & .2549     & .2473 & .2370   \\
                & Quora   & .0363  & .0143  & .0894     & .0182 & .1017   \\
                & SNLI    & .2198  & .2566  & .3158     & .2517 & .2938   \\
                & IMDb    & .2014  & .2188  & .2494     & .2209 & .2309   \\
                & SST-2   & .1326  & .1093  & .1372     & .1101 & .1400   \\
            \hhline{---~~~~}
            \multirow{5}{*}{\rot{LIME}}
                & MNLI    &        & .3281   & .2444    & .3187 & .2269   \\
                & Quora   &        &  .2099  & .1900    & .2037 & .1670   \\
                & SNLI    &        & .2673   & .1676    & .2481 & .1566   \\
                & IMDb    &        & .6538   & .5854    & .6486 & .5584   \\
                & SST-2   &        & .4968   & .4734    & .4962 & .4422   \\
            \hhline{----} 
            \multirow{5}{*}{\rot{Int-Grad}}
                & MNLI   &    &     & .4984   & .8138 lightgray & .4021   \\
                & Quora  &    &     & .2906   & .7420 lightgray & .2290   \\
                & SNLI   &    &     & .2461   & .6535 lightgray & .2165   \\
                & IMDb   &    &     & .7331   & .9409 lightgray & .6994   \\
                & SST-2  &    &     & .8683   & .9707 lightgray & .8063   \\
            \hhline{-----} 
            \multirow{5}{*}{\rot{DeepLIFT}}
                & MNLI    &        &&     &   .4987    & .6208 lightgray  \\
                & Quora   &        &&     &   .3158    & .6179 lightgray  \\
                & SNLI    &        &&     &   .2557    & .5791 lightgray  \\
                & IMDb    &        &&     &   .7378    & .8593 lightgray  \\
                & SST-2   &        &&     &   .8682    & .8729 lightgray  \\
            \hhline{------}
            \multirow{5}{*}{\rot{Grad-SHAP}}
                & MNLI    &        &         &     &       & .4015   \\
                & Quora   &        &         &     &       & .2433   \\
                & SNLI    &        &         &     &       & .2219   \\
                & IMDb    &        &         &     &       & .7021   \\
                & SST-2   &        &         &     &       & .8056   \\
            \hhline{-------}
            \end{tabular}}%
        \caption{BiLSTM}\label{tab:bilstm_results}%
    \end{subtable}
    
    \begin{subtable}[b]{0.97\columnwidth}%
        \renewcommand{\arraystretch}{1.1}%
        \resizebox{\columnwidth}{!}{%
        \begin{tabular}{clRRRRR}
            \toprule
                &  & \multicolumn{1}{c}{LIME} & \multicolumn{1}{c}{Int-Grad} & \multicolumn{1}{c}{DeepLIFT} & \multicolumn{1}{c}{Grad-SHAP} & \multicolumn{1}{c}{Deep-SHAP}            \\
            \hhline{-------}\addlinespace[0.1mm]
            \multirow{5}{*}{\rot{Attn Roll}}
                & MNLI    & .2678  & .1891  & .2432     & .1905 & .2067    \\
                & Quora   & .1622  & .0574  & .2267     & .0518 & .2257   \\
                & SNLI    & .1434  & .1645  & .2214     & .1600 & .1796   \\
                & IMDb    & .1259  & .1818  & .2516     & .1432 & .2303   \\
                & SST-2   & .1359  & .0511  & .1328     & .0737 & .1291   \\
            \hhline{---~~~~}
            \multirow{5}{*}{\rot{LIME}}
                & MNLI    &        & .1794   & .1526    & .1592 & .1205 \\
                & Quora   &        & .1407   & .0032    & .1144 & .0095   \\
                & SNLI    &        & .1529   & .0925    & .1104 & .0593   \\
                & IMDb    &        & .1050   & .0696    & .0929 & .0655   \\
                & SST-2   &        & .2861   & .0618    & .2414 & .0499   \\
            \hhline{----} 
            \multirow{5}{*}{\rot{Int-Grad}}
                & MNLI    &        && .2153    & .4780 lightgray & .1708   \\
                & Quora   &        && .0625    & .4674 lightgray & .0529   \\
                & SNLI    &        && .0955    & .3932 lightgray & .0700   \\
                & IMDb    &        && .1433    & .5495 lightgray & .1246   \\
                & SST-2   &        && .0498    & .4987 lightgray & .0381   \\
            \hhline{-----} 
            \multirow{5}{*}{\rot{DeepLIFT}}
                & MNLI    &        &&     &   .2324   & .4985 lightgray  \\
                & Quora   &        &&     &   .0637   & .5951 lightgray  \\
                & SNLI    &        &&     &   .1181   & .5554 lightgray  \\
                & IMDb    &        &&     &   .1306   & .4830 lightgray  \\
                & SST-2   &        &&     &   .0522   & .4514 lightgray  \\
            \hhline{------} 
            \multirow{5}{*}{\rot{Grad-SHAP}}
                & MNLI    &        &         &     &       & .1752   \\
                & Quora   &        &         &     &       & .0535   \\
                & SNLI    &        &         &     &       & .0851   \\
                & IMDb    &        &         &     &       & .1093   \\
                & SST-2   &        &         &     &       & .0419   \\
            \hhline{-------} 
        \end{tabular}}%
        \caption{DistilBERT}\label{tab:distilbert_results}%
    \end{subtable}%
\end{table}%

\subsection{\ac{XAI} methods rarely correlate with one another}
Table \ref{tab:results} displays the Kendall-$\tau$ correlations for: (a) the BiLSTM model, and (b) the DistilBERT model.
Since the agreement between \ac{XAI} methods and their SHAP approximations is biased by algorithmic similarity, we do not include their comparisons in our calculations of average agreement.
We answer \textbf{RQ(i)} and \textbf{RQ(ii)} by showing our \ac{XAI} methods neither agree with each other ($\text{mean}=0.2684$) nor with attention-based explanations ($\text{mean}=0.1736$) across models and tasks.

\subsection{Correlation is model and task dependent}
For \textbf{RQ(a)}, the agreement between non-attention-based XAI methods is lower for DistilBERT ($\text{mean}=0.1088$) than for the BiLSTM ($\text{mean}=0.4281$). 
Average agreement between the XAI methods and attention-based explanations is comparable for both models ($\text{DistilBERT mean}=0.1658$, $\text{BiLSTM mean}=0.1814$).  
Regarding \textbf{RQ(b)}, the total agreement across all methods is higher for the single-sequence datasets ($\text{combined model mean}=0.273$) than for the pair-sequence datasets ($\text{combined model mean}=0.1883$).
This difference is particularly noticeable for the BiLSTM ($\text{single-sequence mean}=0.4219$, $\text{pair-sequence mean}=0.2308$).

\section{Discussion \& Conclusion} \label{sec:conclusion}


The \emph{agreement as evaluation} paradigm assumes — at least implicitly — the desirability of an XAI method decreases monotonically with its correlation to some unobserved `ideal’. However, there are reasons to doubt whether this assumption holds. For instance, input rankings may only capture a narrow slice of the model's behavior such that many equally faithful compressions exist. And, since many tasks may be too complex for humans to judge token-level importance, there may also be many plausible rankings. While a handful of highly polar tokens are generally indicative of the class label in binary sentiment classification \cite{sun-lu-2020-understanding}, annotators may be unsure how to rank the other tokens. The problem only gets worse in the pair-sequence setting. For example, if two words indicate a contradiction, which one is more important?  There is a reason that rationale collections are normally limited to binary relevance labels or free-form explanations\footnote{See \cite{wiegreffe2021teach} and \cite{deyoung-etal-2020-eraser} for good reviews of explainability datasets in NLP.}. Thus, when agreement is measured in the presence of multiple faithful and plausible rankings, XAI methods will look deceptively problematic.


We observe low agreement among \ac{XAI} methods when explaining more complex models and tasks. If we embraced \emph{agreement as evaluation}, we would be obligated to conclude at most one of our chosen XAI methods is near the ideal; implying the other methods cannot explain the more complex Transformer-based model and pair-sequence tasks. Instead, we interpret our results as evidence against the underlying assumptions of \emph{agreement as evaluation}, and conclude that agreement is not a suitable method of evaluation.

Without an external ground-truth explanation \citep[like those constructed by][]{yalcin2021evaluating}, all rank correlation tells us is whether or not two rankings are similar. For this reason, we recommend practitioners stop using \textit{agreement as evaluation}. Instead, we recommend using robust, theoretically-motivated measures of an \ac{XAI} method's quality, such as those proposed by \citet{atanasova-etal-2020-diagnostic}.


Agreement can still be informative, even if it is unsuitable as an evaluation measure. For example, it may reveal how theoretical properties manifest in practice. While algorithms that approximate Shapley Values are normally referenced with the umbrella term `SHAP', \citet{ethayarajh2021attention} show that \emph{attention flow} \cite{abnar-zuidema-2020-quantifying} is also a Shapley Value explanation. Interestingly, we observe low agreement between Grad-SHAP and Deep-SHAP ($\text{combined model mean} = 0.2839$) and between \emph{attention flow} and our chosen SHAP approximations as well ($\text{mean}=0.1726$, see our supplementary material). As previously argued, this does not mean these methods are wrong, merely that we cannot assume they are interchangeable.


\section*{Acknowledgments}
We thank Maarten de Rijke and the anonymous reviewers for their comments and suggestions.
This research was supported by the Dutch National Police, and the Netherlands Organisation for Scientific Research (NWO) under project nr. 652.001.003. All content represents the opinion of the authors, which is not necessarily shared or endorsed by their respective employers and/or sponsors.

\nocite{langley00}

\bibliography{anthology,references}
\bibliographystyle{icml2021}

\end{document}


\appendix
\section{Attention Flow}
\begin{table}[h!]
    \centering
    \caption{Mean Kendall-$\tau$ between the explanations given by \emph{attention flow} and our chosen XAI methods for the DistilBERT model when applied to 500 instances of the test portion of each dataset. IMDb is not included among these datasets, because the long sequences made the \emph{attention flow} computation unfeasible.}\label{tab:attn_flow_apx}
    \setlength\tabcolsep{2pt}
    \renewcommand{\arraystretch}{1.17}
    \resizebox{0.97\columnwidth}{!}{%
    \begin{tabular}{clRRRRR}
            &  & \multicolumn{1}{c}{LIME} & \multicolumn{1}{c}{Int-Grad} & \multicolumn{1}{c}{DeepLIFT} & \multicolumn{1}{c}{Grad-SHAP} & \multicolumn{1}{c}{Deep-SHAP}            \\
        \hhline{-------}\addlinespace[0.1mm]
        \multirow{4}{*}{\rot{Attn Flow}}
            & MNLI    & .1326  & .1251  & .2159  & .1227  & .2148   \\
            & Quora   & .0853  & .2426  & .0367  & .0241  & .2319   \\
            & SNLI    & .0844  & .0753  & .2178  & .0571  & .2149   \\
            & SST-2   & .1795  & .0689  & .1286  & .0811  & .1202   \\
        \hhline{-------} 
    \end{tabular}}
\end{table}\vspace{-1em}

Despite \emph{attention flow}, Grad-SHAP, and Deep-SHAP all (supposedly) being valid Shapley Value explanations, agreement is low.
\section{Reproducibility Checklist}
In this Appendix, we include information about our experiments from the Reproducibility Checklist.

\subsection{For all reported experimental results}

\subsubsection{A clear description of the mathematical setting, algorithm, and/or model}
We clearly explain our methods in Section 3 and our models, datasets, and experiments in Section 4.

\subsubsection{Submission of a zip file containing source code, with specification of all dependencies, including external libraries, or a link to such resources (while still anonymized)}
Our code is publicly available at 
\url{github.com/sfschouten/court-of-xai}

\subsubsection{Description of computing infrastructure used}
We conducted our experiments on Amazon Web Services \texttt{g4dn.xlarge} EC2 instances using an NVIDIA T4 GPU with 16GB of RAM. The version of PyTorch was \texttt{1.6.0+cu101}.

\subsubsection{Average runtime for each approach}
Refer to Table \ref{tab:runtime} for the average time to train each model on each dataset.

\subsubsection{Number of parameters in each model}
The DistilBERT model contained 66955779 trainable parameters and the BiLSTM model contained 12553519 trainable parameters, as reported by the AllenNLP library.

\subsubsection{Corresponding validation performance for each reported test result}
Table \ref{tab:val_accuracy} details the validation performance of the best model weights for each dataset.

\subsubsection{Explanation of evaluation metrics used, with links to code}
We evaluate our models by their accuracy. We evaluate the correlation (agreement) between XAI methods using Kendall's-$\tau$. Both of these metrics are explained in Section 3. The code is available at the previously listed URL.

\subsection{For all experiments with hyperparameter search}
The items in this part of the Reproducibility Checklist are not applicable to our paper.

\subsection{For all datasets used}

\subsubsection{Relevant statistics such as number of examples}
Table \ref{tab:dataset_statistics} lists the number of instances in each split of each dataset.

\subsubsection{Details of train/validation/test splits}
Split details are outlined in Section 4.1. See below for links to each dataset.

\subsubsection{Explanation of any data that were excluded, and all pre-processing steps}
Details of data exclusion and pre-processing steps are outlined in Section 4.1.

\subsubsection{A link to a downloadable version of the data}
 Links to download versions of all datasets are included in our code repository. For posterity, links to all datasets are listed here: \textbf{SST-2}\footnote{\url{https://github.com/successar/AttentionExplanation/tree/master/preprocess/SST}}, \textbf{IMDb}\footnote{\url{https://github.com/successar/AttentionExplanation/tree/master/preprocess/IMDB}}, \textbf{SNLI}\footnote{\url{https://nlp.stanford.edu/projects/snli/}}, \textbf{MNLI}\footnote{\url{https://cims.nyu.edu/~sbowman/multinli/}}, XNLI\footnote{\url{https://cims.nyu.edu/~sbowman/xnli/}}. Our \textbf{Quora} Question Pair dataset will be made available upon publication.

\subsubsection{For new data collected, a complete description of the data collection process, such as instructions to annotators and methods for quality control}
We did not collect new data for this paper.

\begin{table}
    \centering
    \begin{tabular}{crr}
     \toprule
        & BiLSTM & DistilBERT \\
        \midrule
        MNLI    & 8.65 $\pm$ 0.635 & 296.228 $\pm$ 48.859 \\
        Quora   & 7.567 $\pm$ 1.404 & 380.056 $\pm$ 124.911 \\
        SNLI    & 31.495 $\pm$ 5.618 & 126.395 $\pm$ 22.909 \\
        IMDb    & 1.122 $\pm$ 0.107 & 24.2 $\pm$ 1.212 \\
        SST-2   & 0.216 $\pm$ 0.029 & 2.833 $\pm$ 0.65 \\
    \bottomrule
    \end{tabular}
    \caption{Number of minutes (average $\pm$ standard deviation) required to train each model on each dataset reported across three seeds.}
    \label{tab:runtime}
\end{table}

\begin{table}
    \centering
    \begin{tabular}{ccc}
     \toprule
        & BiLSTM & DistilBERT \\
        \midrule
        MNLI    & 67.088 $\pm$ 0.190 & 77.338 $\pm$ 0.251 \\
        Quora   & 83.232 $\pm$ 0.139 & 88.801 $\pm$ 0.055 \\
        SNLI    & 81.535 $\pm$ 0.041 & 87.679 $\pm$ 0.075 \\
        IMDb    & 87.975 $\pm$ 1.375 & 88.587 $\pm$ 0.489 \\
        SST-2   & 80.696 $\pm$ 0.403 & 83.066 $\pm$ 0.692 \\
    \bottomrule
    \end{tabular}
    \caption{Validation accuracy (average $\pm$ standard deviation) of the selected model epoch reported across three seeds.}
    \label{tab:val_accuracy}
\end{table}

\begin{table}
    \centering
    \begin{tabular}{crrr}
     \toprule
        & Training & Validation & Test \\
        \midrule
        MNLI    & 392702 & 10000 & 5000 \\
        Quora   & 323426 & 40429 & 40431 \\
        SNLI    & 550152 & 10000 & 10000 \\
        IMDb    & 17212 & 4304 & 4363 \\
        SST-2   & 8544 & 1101 & 2210 \\
    \bottomrule
    \end{tabular}
    \caption{Number of instances in each split of each dataset before any exclusions based on length (see Section 4.1). Since MultiNLI has no publicly available test set, we use the English subset of the XNLI dataset.}
    \label{tab:dataset_statistics}
\end{table}